\documentclass{article}
\usepackage{float}

\usepackage[english]{babel}

\usepackage[letterpaper,top=2cm,bottom=2cm,left=3cm,right=3cm,marginparwidth=1.75cm]{geometry}

\usepackage{amsmath}
\usepackage{graphicx}
\usepackage[colorlinks=true, allcolors=blue]{hyperref}

\PassOptionsToPackage{unicode}{hyperref}
\PassOptionsToPackage{hyphens}{url}

\usepackage{xcolor}
\usepackage{amsmath,amssymb}

\usepackage[T1]{fontenc}
\usepackage[utf8]{inputenc}
\usepackage{textcomp}
\usepackage{lmodern}
\ifPDFTeX\else
\fi
\IfFileExists{upquote.sty}{\usepackage{upquote}}{}
\IfFileExists{microtype.sty}{
  \usepackage[]{microtype}
  \UseMicrotypeSet[protrusion]{basicmath} 
}{}

\usepackage{longtable,booktabs,array}
\usepackage{multirow}
\usepackage{calc} 
\usepackage{etoolbox}
\makeatletter
\patchcmd\longtable{\par}{\if@noskipsec\mbox{}\fi\par}{}{}
\makeatother
\IfFileExists{footnotehyper.sty}{\usepackage{footnotehyper}}{\usepackage{footnote}}
\makesavenoteenv{longtable}
\usepackage{graphicx}
\makeatletter
\newsavebox\pandoc@box
\newcommand*\pandocbounded[1]{
  \sbox\pandoc@box{#1}%
  \Gscale@div\@tempa{\textheight}{\dimexpr\ht\pandoc@box+\dp\pandoc@box\relax}%
  \Gscale@div\@tempb{\linewidth}{\wd\pandoc@box}%
  \ifdim\@tempb\p@<\@tempa\p@\let\@tempa\@tempb\fi
  \ifdim\@tempa\p@<\p@\scalebox{\@tempa}{\usebox\pandoc@box}%
  \else\usebox{\pandoc@box}%
  \fi%
}
\makeatother
\usepackage{bookmark}
\IfFileExists{xurl.sty}{\usepackage{xurl}}{} 
\hypersetup{
  hidelinks,
  pdfcreator={LaTeX via pandoc}}
\usepackage{indentfirst}

\author{}
\date{}

\title{STA-Net: A Decoupled Shape and Texture Attention Network for Lightweight Plant Disease Classification\thanks{This work is supported by the Innovation Training Program for College Students in Sichuan Province (No.202410626002).}}
\usepackage{authblk}

\author[1]{Zongsen Qiu}
\author[2]{Jianjun Wang\thanks{Corresponding author. Email: wangjianjun02@163.com}}
\author[2]{Yue Zhou}
\author[2]{Zibo Zhou}
\author[2]{Rui Chen}

\affil[1]{College of Information Engineering, Sichuan Agricultural University, YaAn, China}
\affil[2]{College of Science, Sichuan Agricultural University, YaAn, China}

\begin{document}
\maketitle

\begin{abstract}
Responding to rising global food security needs, precision agriculture and deep learning-based plant disease diagnosis have become crucial. Yet, deploying high-precision models on edge devices is challenging. Most lightweight networks use attention mechanisms designed for generic object recognition, which poorly capture subtle pathological features like irregular lesion shapes and complex textures. To overcome this, we propose a twofold solution: first, using a training-free neural architecture search method (DeepMAD) to create an efficient network backbone for edge devices; second, introducing the Shape-Texture Attention Module (STAM). STAM splits attention into two branches—one using deformable convolutions (DCNv4) for shape awareness and the other using a Gabor filter bank for texture awareness. On the public CCMT plant disease dataset, our STA-Net model (with 401K parameters and 51.1M FLOPs) reached 89.00\% accuracy and an F1 score of 88.96\%. Ablation studies confirm STAM significantly improves performance over baseline and standard attention models. Integrating domain knowledge via decoupled attention thus presents a promising path for edge-deployed precision agriculture AI. The source code is available at \url{https://github.com/RzMY/STA-Net}.
\end{abstract}

\section{Introduction}
Ensuring global food security amid escalating population pressures stands as one of the foremost challenges of the 21st century. Precision agriculture—an essential technological paradigm for addressing this challenge—seeks to harness information technology to optimize agricultural production, thereby enhancing crop yields and minimizing resource wastage. Among its pivotal components, the rapid and accurate diagnosis of plant diseases is critical for reducing crop losses, optimizing pesticide application, and promoting agricultural sustainability. In recent years, convolutional neural networks (CNNs), a representative deep learning technique, have achieved significant breakthroughs in image recognition, with their application in plant disease classification demonstrating considerable promise. However, although large-scale network models such as ResNet~\cite{he2016deep} and VGG~\cite{simonyan2014very} attain high classification accuracy, their substantial computational requirements and parameter counts restrict their deployment on resource-constrained mobile or embedded devices (i.e., edge devices), thus challenging their suitability for real-time field diagnostics. Consequently, the exploration of lightweight neural networks tailored for edge computing has emerged as a vital research direction in this domain.

However, current research on lightweight models is generally limited by a "one-size-fits-all" approach. Mainstream lightweight architectures, such as MobileNets and ShuffleNets, along with their associated attention mechanisms—for example, Squeeze-and-Excitation (SE) and CBAM—are predominantly designed for general visual recognition tasks like ImageNet, which emphasize distinguishing objects with markedly different characteristics. In contrast, plant disease identification is essentially a fine-grained visual classification (FGVC) task~\cite{zhao2017diversified}, where the primary challenge is to capture the subtle visual variations among diseases within the same species. The key discriminative features in this context manifest in two aspects: first, the irregular geometric shapes of lesions; and second, the unique surface textures present in disease-affected regions (e.g., downy mildew or rust spots). As general-purpose attention mechanisms do not explicitly model these specific pathological features, they are inefficient at capturing critical discriminative information, thereby limiting the potential for enhanced model performance.

To overcome the limitations of existing research, this study proposes a novel approach specifically tailored for plant disease recognition tasks. The proposed method makes advances on two fronts. First, at the network architecture level, we employ an efficient, training-free neural architecture search (NAS) method called DeepMAD. Operating under predefined computational resource constraints—such as parameter count and FLOPs—DeepMAD rapidly generates lightweight backbone networks that are optimized for hardware efficiency. Second, at the attention mechanism level, we introduce the Shape-Texture Attention Module (STAM), a core innovation of this research. This module decouples the process of spatial attention learning into two distinct branches: one that focuses on shape perception to delineate lesion contours and another that emphasizes texture perception to identify pathological patterns. By integrating prior knowledge of plant disease visual features into the network architecture, this design effectively guides the model in efficiently learning critical discriminative information.

The primary contributions of this work are summarized as follows:
\begin{enumerate}
    \item We propose a novel lightweight spatial attention module, STAM, which employs a decoupled design to integrate deformable convolution and Gabor filters for independently capturing the shape and texture features of plant diseases. This module is specifically optimized for fine-grained visual classification tasks.
    \item We develop a comprehensive and efficient model construction workflow. First, an efficient backbone network is generated using a training-free neural architecture search (NAS) method. Subsequently, the proposed domain-specific attention module (STAM) is integrated into the network.
    \item We conduct extensive experiments on the publicly available CCMT plant disease dataset. The results indicate that our model outperforms the baseline and models employing general attention mechanisms while maintaining a low computational cost.
    \item To promote further research in this domain, we will publicly release the code, pre-trained weights, and usage instructions, thereby ensuring the reproducibility of our research findings.
\end{enumerate}

\section{Related Work}

\subsection{Lightweight CNN Architecture}
To address the challenge of deploying deep learning models on devices with limited computational resources, researchers have proposed a variety of efficient, lightweight CNN architectures. The MobileNet series, in particular, has emerged as a benchmark solution. MobileNetV1~\cite{howard2017mobilenets} significantly reduces computational complexity by substituting standard convolutions with depthwise separable convolutions. Building on this foundation, MobileNetV2~\cite{sandler2018mobilenetv2} introduced an inverted residual structure that further enhances model performance and efficiency through an "expansion–depthwise convolution–projection" scheme. MobileNetV3~\cite{howard2019searching} subsequently integrated neural architecture search (NAS) with the h-swish activation function and the Squeeze-and-Excitation (SE) module~\cite{hu2018squeeze} to achieve further improvements in efficiency and accuracy. In addition, other lightweight networks, such as ShuffleNet~\cite{zhang2018shufflenet}—which utilizes channel shuffle operations—and EfficientNet~\cite{tan2019efficientnet}—which scales model depth, width, and resolution in a composite manner—have demonstrated notable advantages in this domain.

These contributions offer valuable building blocks and design principles for constructing efficient convolutional neural networks (CNNs). However, these architectures are generally designed for broad applicability, and their configurations (e.g., number of channels, expansion ratios) may not be optimal for specialized tasks, such as plant disease identification. Motivated by the effectiveness of the inverted residual blocks in MobileNetV2 and MobileNetV3, this study employs neural architecture search (NAS) techniques to tailor a design specifically for plant disease classification, thereby achieving an improved balance between accuracy and efficiency.

\subsection{Neural Architecture Search (NAS)}
Neural Architecture Search (NAS) seeks to automate the design of high-performance network architectures by replacing the traditional, expert-driven, and labor-intensive manual design process. Early NAS approaches, such as those leveraging reinforcement learning~\cite{zoph2016neural} or evolutionary algorithms~\cite{real2017large}, succeeded in discovering effective architectures but typically required thousands of GPU hours, rendering them prohibitively expensive. Subsequent methodologies, including the DARTS framework~\cite{liu2018darts}, introduced differentiable strategies that significantly reduced search time, although they still necessitated extensive training.

In recent years, training-free NAS~\cite{chen2021neural} has emerged as a prominent research direction. These approaches predict a network's final performance by evaluating proxy metrics without any model training. For instance, some methods assess the network's data processing capabilities at initialization or exploit its gradient information as proxy metrics. The DeepMAD method~\cite{chen2023deepmad} employed in this study belongs to this category; grounded in entropy theory, it rapidly evaluates and filters network architectures on a CPU, thereby enabling the selection of optimal candidate architectures under specified resource constraints.

In this research, DeepMAD is utilized as a robust tool to achieve our objectives rather than as the central innovation. We consider it a practical and efficient means to swiftly generate a lightweight backbone network that is compatible with the STAM module and optimized for edge deployment. This pragmatic choice allows us to concentrate our efforts on domain-specific innovations in attention mechanisms.

\subsection{Attention Mechanism in Computer Vision}
The attention mechanism, inspired by the dynamic nature of the human visual system, enables models to precisely focus on essential components within the input data. Today, it constitutes an integral module in state-of-the-art convolutional neural networks. Functionally, the attention mechanism can be broadly classified into channel attention and spatial attention. Channel attention primarily quantifies the importance of different feature channels. In this domain, the Squeeze-and-Excitation (SE) network~\cite{hu2018squeeze} represents a milestone; by integrating global average pooling (squeeze) with a compact multi-layer perceptron (excitation), it adaptively learns channel weights and dynamically recalibrates channel responses. Subsequent approaches, such as the Efficient Channel Attention (ECA) method~\cite{wang2020eca}, have further refined the excitation design to enhance performance.

In contrast, spatial attention concentrates on identifying key regions within the feature map. For instance, the Convolutional Block Attention Module (CBAM)~\cite{woo2018cbam} extends SE by incorporating a dedicated spatial attention branch that applies pooling operations along the channel dimension and utilizes convolutional layers to generate a spatial attention map, thereby further refining spatial feature representations.

However, irrespective of whether attention is applied at the channel or spatial level, prevailing mechanisms typically maintain a high degree of generality. These methods predominantly learn to determine “what” or “where” to focus solely from data, without incorporating task-specific prior knowledge into their architectural design. In fine-grained visual categorization (FGVC) tasks, such as plant disease identification, this generic design can result in inefficient resource utilization because the model must learn from scratch how to attend to irregular shapes and distinctive textures without any structured preliminary guidance.

The proposed STAM module addresses this limitation by explicitly integrating domain knowledge into the attention mechanism. Specifically, for plant disease identification, we posit that shape and texture represent two critical, complementary visual dimensions. Accordingly, dedicated processing branches are designed for each dimension: DCNv4~\cite{wang2024dcnv4} is employed to manage deformable shapes, while Gabor filters~\cite{jain1991unsupervised} are utilized to extract texture features. This integration of a strong inductive bias facilitates the model’s focus on the most salient features for the task, ultimately enhancing both efficiency and accuracy.

\section{Proposed Method}
\subsection{Overall Framework}
\begin{figure}[h]
\centering
\includegraphics[width=\linewidth]{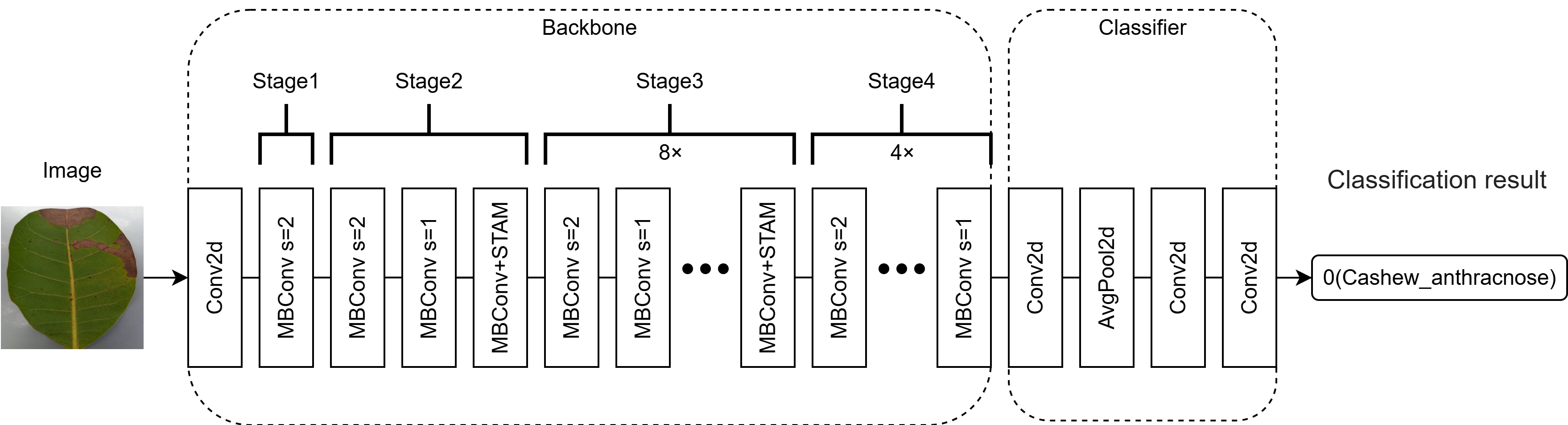}
\caption{\label{fig:STA-Net}Overall Architecture of STA-Net}
\end{figure}

The proposed STA-Net framework is designed to balance high accuracy with efficiency, making it particularly well-suited for deployment on edge devices. Its core design principle, encapsulated by the "efficient backbone + precise attention" paradigm, is realized by integrating a lightweight convolutional neural network backbone—generated using a training-free NAS method (DeepMAD)—with strategically embedded STAM modules. As illustrated in Figure \ref{fig:STA-Net}, the overall architecture includes a backbone that performs layered feature extraction, transforming low-level input features into high-dimensional semantic representations. Concurrently, the STAM modules refine the spatial information in the intermediate feature maps, enhancing the network’s ability to identify critical pathological regions. The refined features are then propagated through subsequent layers, ultimately leading to a classification head that outputs the prediction results.

\subsection{Lightweight network architecture}
We employ DeepMAD for designing the backbone network due to its training-free nature and robust mathematical underpinnings. This approach circumvents the high computational costs typically associated with traditional neural architecture search (NAS) by utilizing an entropy proxy indicator to evaluate network structures. It efficiently identifies the optimal combination of architectural hyperparameters—subject to specific resource constraints (e.g., a total parameter count below 300k)—from a vast search space, eliminating the need for any training operations. These hyperparameters include kernel sizes, input channel counts, expansion ratios, output channel counts, strides, and repetition counts within each inverted residual block.

\begin{figure}[H]
\centering
\includegraphics[width=\linewidth]{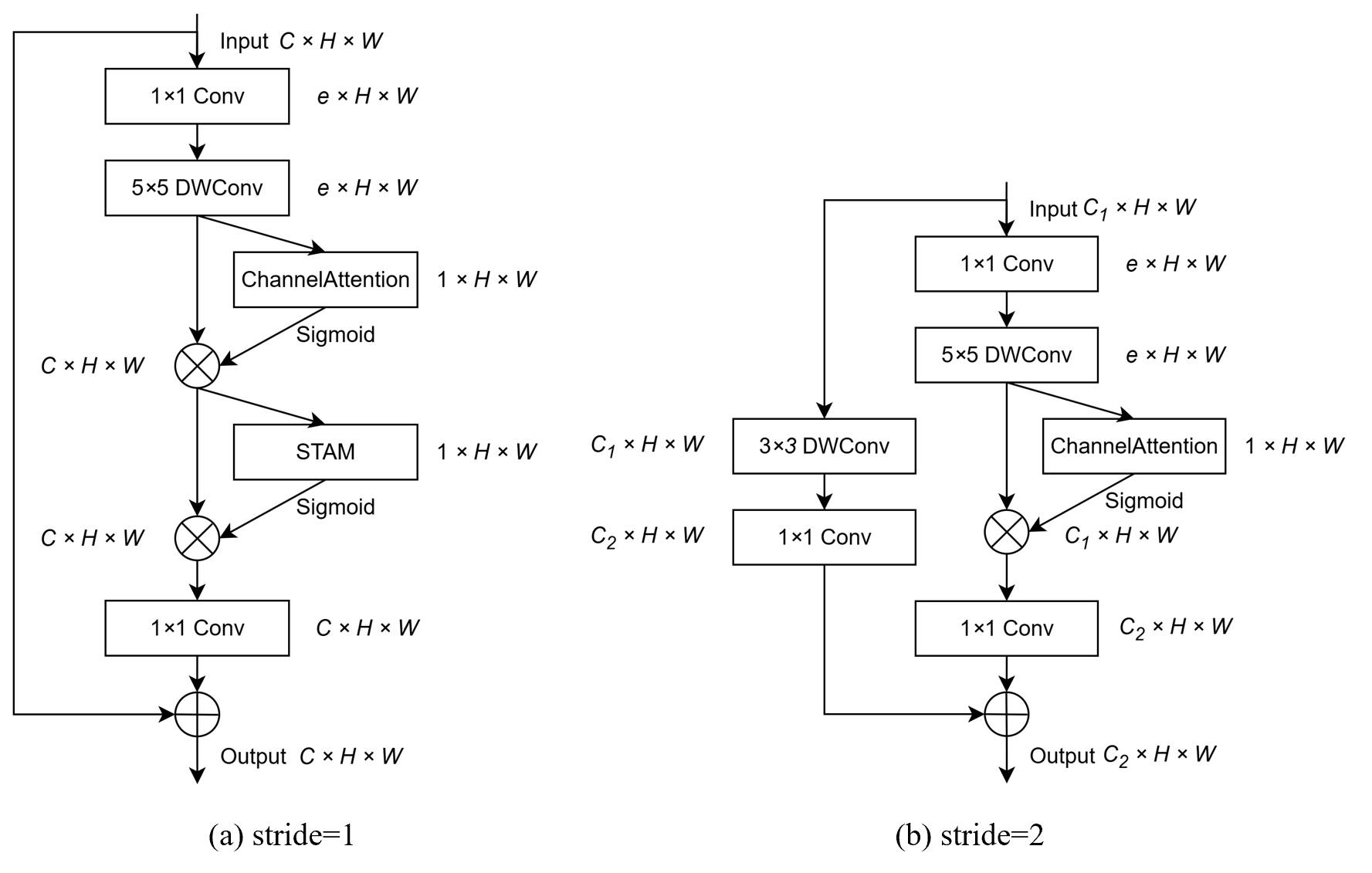}
\caption{\label{fig:MBConv}MBConv building blocks under both stride=1 and stride=2 conditions}
\end{figure}

In this study, we employ the MBConv module, as utilized in MobileNetV3~\cite{howard2019searching}, as the fundamental network building block. Figure \ref{fig:MBConv} illustrates the detailed architecture of the MBConv block, which comprises a 1×1 expansion convolution, a k×k depthwise separable convolution, an optional combination of channel and spatial attention mechanisms, and a 1×1 compression convolution, all complemented by a residual connection. For downsampling blocks with a stride of 2, the residual connection is subjected to additional processing and typically excludes the spatial attention mechanism. This design not only facilitates efficient feature representation but also significantly reduces both the parameter count and computational overhead. By integrating the MBConv blocks and constructing the backbone network based on hyperparameters derived through a search process, we subsequently design a classification head, culminating in the formation of the network designated as STA-Net. The detailed configuration of each layer is presented in Table \ref{tab:sta}.

\begin{longtable}[]{@{}
  >{\centering\arraybackslash}p{(\linewidth - 14\tabcolsep) * \real{0.1422}}
  >{\centering\arraybackslash}p{(\linewidth - 14\tabcolsep) * \real{0.2398}}
  >{\centering\arraybackslash}p{(\linewidth - 14\tabcolsep) * \real{0.1548}}
  >{\centering\arraybackslash}p{(\linewidth - 14\tabcolsep) * \real{0.0931}}
  >{\centering\arraybackslash}p{(\linewidth - 14\tabcolsep) * \real{0.1053}}
  >{\centering\arraybackslash}p{(\linewidth - 14\tabcolsep) * \real{0.1295}}
  >{\centering\arraybackslash}p{(\linewidth - 14\tabcolsep) * \real{0.0820}}
  >{\centering\arraybackslash}p{(\linewidth - 14\tabcolsep) * \real{0.0532}}@{}}
\caption{\label{tab:sta}Specification for STA-Net. SE denotes whether there is a Squeeze-and-Excitation in that block. STAM denotes whether there is a STAM in that block. NL denotes the type of nonlinearity used. Here, HS denotes Hardswish and RE denotes ReLU. s denotes stride.}\\
\toprule\noalign{}
\begin{minipage}[b]{\linewidth}\centering
Input
\end{minipage} & \begin{minipage}[b]{\linewidth}\centering
Operator
\end{minipage} & \begin{minipage}[b]{\linewidth}\centering
exp size
\end{minipage} & \begin{minipage}[b]{\linewidth}\centering
out
\end{minipage} & \begin{minipage}[b]{\linewidth}\centering
SE
\end{minipage} & \begin{minipage}[b]{\linewidth}\centering
STAM
\end{minipage} & \begin{minipage}[b]{\linewidth}\centering
NL
\end{minipage} & \begin{minipage}[b]{\linewidth}\centering
s
\end{minipage} \\
\midrule\noalign{}
\endhead
\bottomrule\noalign{}
\endlastfoot
$224^2 \times 3$ & conv2d, 3×3 & - & 8 & - & - & HS & 2 \\
$112^2 \times 8$ & MBConv, 5×5 & 16 & 8 & True & - & RE & 2 \\
$56^2 \times 8$ & MBConv, 5×5 & 32 & 24 & True & - & RE & 2 \\
$28^2 \times 24$ & MBConv, 5×5 & 32 & 24 & True & - & RE & 1 \\
$28^2 \times 24$ & MBConv, 5×5 & 32 & 24 & True & True & RE & 1 \\
$28^2 \times 24$ & MBConv, 5×5 & 80 & 56 & True & - & HS & 2 \\
$14^2 \times 56$ & MBConv, 5×5 & 80 & 56 & True & - & HS & 1 \\
$14^2 \times 56$ & MBConv, 5×5 & 80 & 56 & True & - & HS & 1 \\
$14^2 \times 56$ & MBConv, 5×5 & 80 & 56 & True & - & HS & 1 \\
$14^2 \times 56$ & MBConv, 5×5 & 80 & 56 & True & - & HS & 1 \\
$14^2 \times 56$ & MBConv, 5×5 & 80 & 56 & True & - & HS & 1 \\
$14^2 \times 56$ & MBConv, 5×5 & 80 & 56 & True & - & HS & 1 \\
$14^2 \times 56$ & MBConv, 5×5 & 80 & 56 & True & True & HS & 1 \\
$14^2 \times 56$ & MBConv, 5×5 & 144 & 72 & True & - & HS & 2 \\
$7^2 \times 72$ & MBConv, 5×5 & 144 & 72 & True & - & HS & 1 \\
$7^2 \times 72$ & MBConv, 5×5 & 144 & 72 & True & - & HS & 1 \\
$7^2 \times 72$ & MBConv, 5×5 & 144 & 72 & True & - & HS & 1 \\
$7^2 \times 72$ & conv2d, 1×1 & - & 512 & - & - & HS & 1 \\
$7^2 \times 512$ & pool, 7×7 & - & - & - & - & - & - \\
$1^2 \times 512$ & conv2d, 1×1 & - & 128 & - & - & HS & 1 \\
$1^2 \times 128$ & conv2d, 1×1 & - & k & - & - & - & 1 \\
\end{longtable}

\subsection{Shape-Texture Attention Module (STAM)}
The STAM module represents the primary innovation of this study. Inspired by the diagnostic approach of plant pathologists—who first localize lesion areas and subsequently scrutinize their internal textures—we introduce the concept of "Perceptual Decoupling." This design strategy decomposes the complex spatial attention task into two more focused subtasks, each managed by a dedicated branch and later integrated through intelligent fusion. This divide-and-conquer approach enables each branch to specialize in learning specific visual patterns, thereby substantially enhancing both the overall attention mechanism and the final classification performance.

\begin{figure}[h]
\centering
\includegraphics[width=0.4\linewidth]{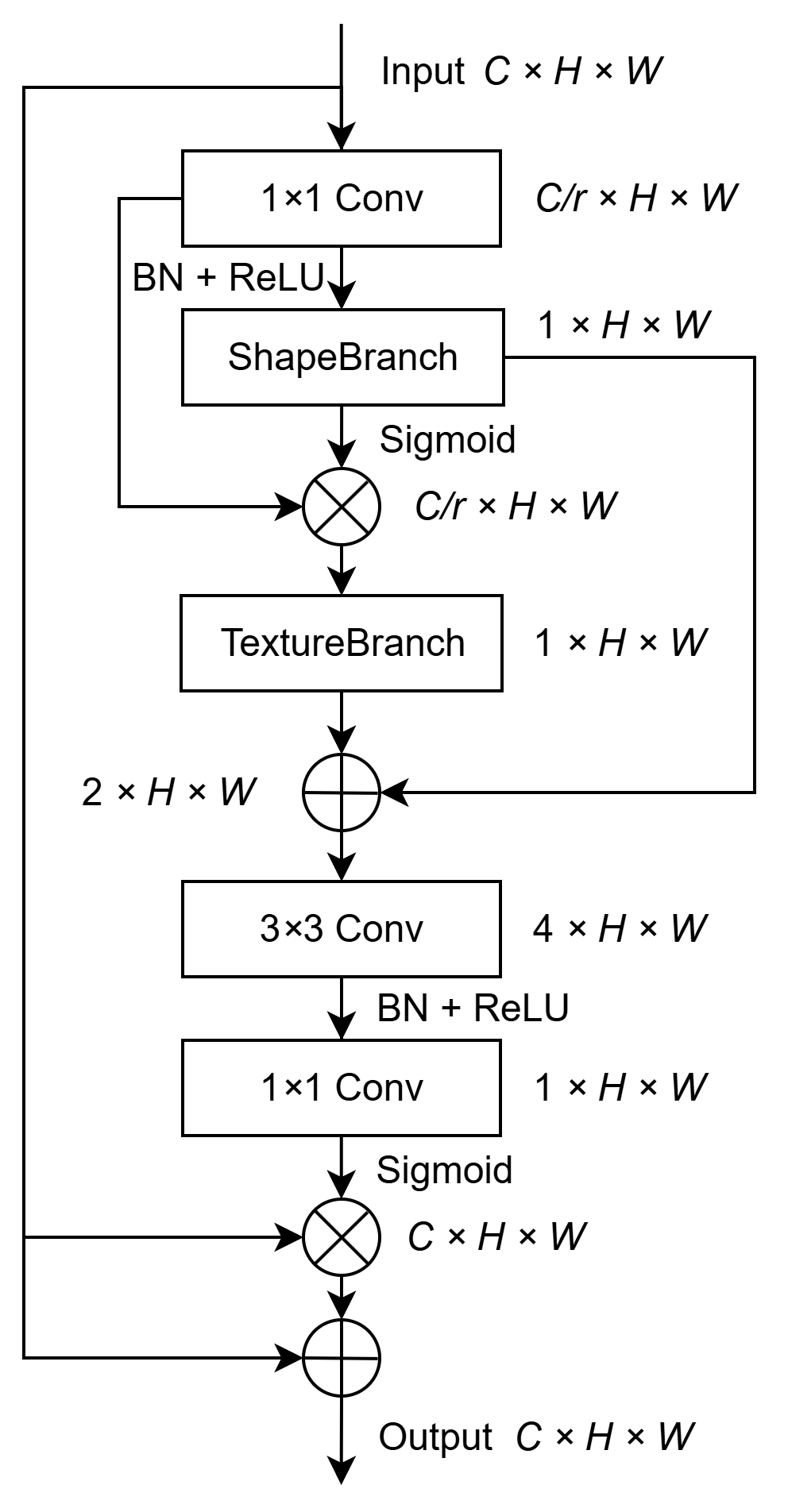}
\caption{\label{fig:STAM}STAM}
\end{figure}

Figure \ref{fig:STAM} depicts the detailed structure of the STAM module. Initially,
the input feature map \(x \in \mathbb{R}^{C \times H \times W}\) is
processed through a 1×1 convolution to generate a channel-compressed
descriptor \(x_{desc} \in \mathbb{R}^{C/r \times H \times W}\), with the
compression degree modulated by the hyperparameter r. A smaller r
results in less compression and improved module performance, albeit at
the expense of increased parameters and computational cost. The
descriptor \(x_{desc}\) is subsequently input into the shape-aware
branch to produce a shape attention map
\(M_{shape} \in \mathbb{R}^{1 \times H \times W}\), which is then
activated with a sigmoid function. The activated \(M_{shape}\) weights
\(x_{desc}\), yielding a shape attention-enhanced feature map that
serves as the input
\(x_{texture} \in \mathbb{R}^{C/r \times H \times W}\) for the
texture-aware branch. The texture-aware branch subsequently generates a
texture attention map
\(M_{texture} \in \mathbb{R}^{1 \times H \times W}\), which is
concatenated with the shape attention map \(M_{shape}\) along the
channel dimension. The fused information is integrated through a small
convolutional module to produce an aggregated attention map
\(M_{stam} \in \mathbb{R}^{1 \times H \times W}\). Finally, after
sigmoid activation, the aggregated attention map is multiplied, in a
residual manner, with the original input x. The overall attention
mechanism is formally summarized by the following mathematical
expression:

\[x_{desc} = f^{1 \times 1}(x)\]
\[M_{shape} = f^{shape}(x_{desc})\]
\[x_{texture} = x_{desc} \otimes \sigma(M_{shape})\]
\[M_{texture} = f^{texture}(x_{texture})\]
\[M_{stam} = \ \sigma(f^{fusion}(\lbrack M_{shape};M_{texture}\rbrack))\]

Among these, \(f^{1 \times 1}\) denotes the 1×1 convolution operation,
whereas \(f^{shape}\) and \(f^{texture}\) represent the shape-aware and
texture-aware branches, respectively. The operator \(\otimes\) indicates
element-wise multiplication, while \(\sigma\) designates the sigmoid
function. Furthermore, \(f^{fusion}\) corresponds to the fusion module,
which comprises a 3×3 convolutional block for increasing dimensionality
and a 1×1 convolutional block for reducing dimensionality. The operation
{[};{]} signifies channel-wise concatenation.

\subsubsection{Shape-Aware Branch (ShapeBranch)}

\begin{figure}[h]
\centering
\includegraphics[width=0.9\linewidth]{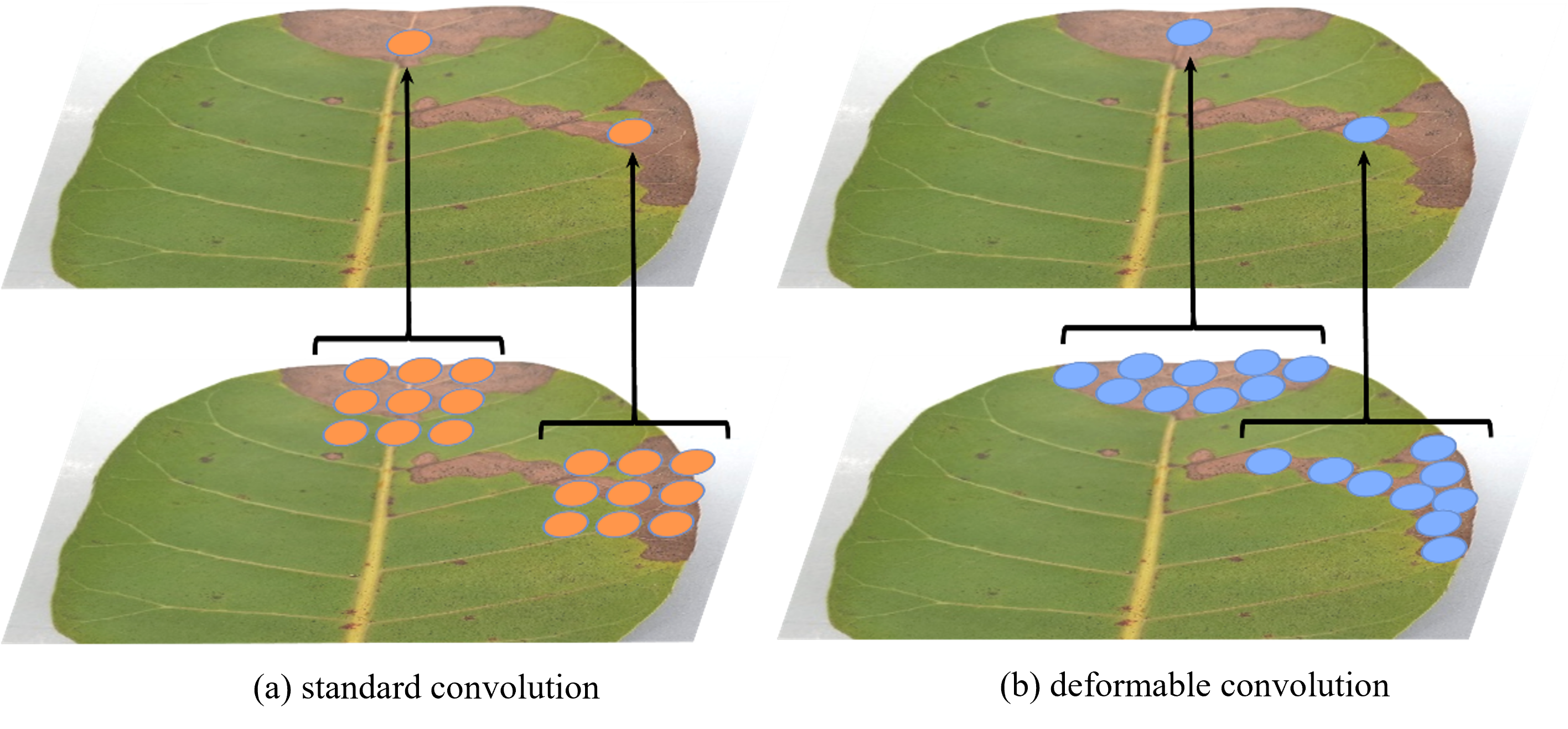}
\caption{\label{fig:Difference}Difference between deformable and standard convolutions}
\end{figure}

\begin{figure}[h]
\centering
\includegraphics[width=0.4\linewidth]{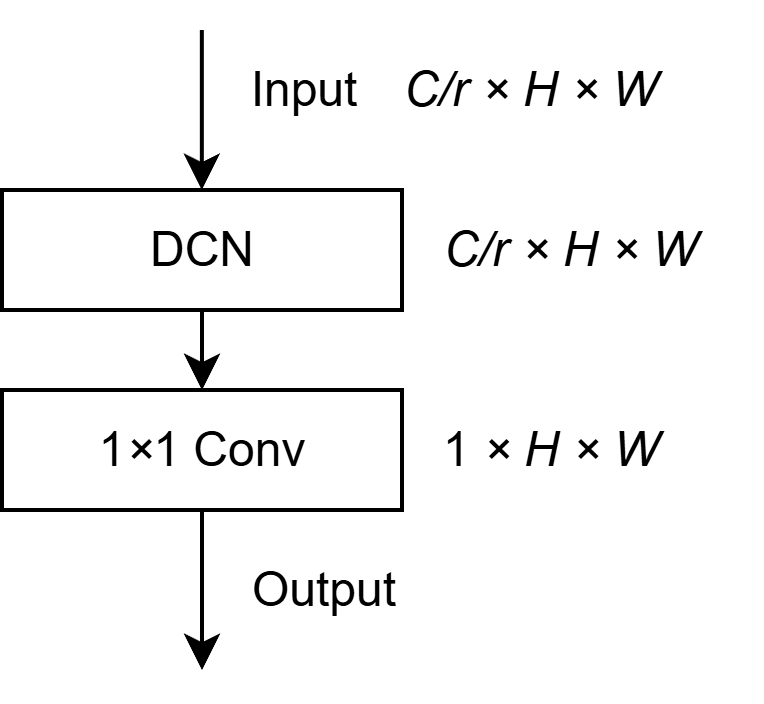}
\caption{\label{fig:Shape}ShapeBranch}
\end{figure}

The morphological characteristics of crop disease spots typically exhibit non-rigid and irregular features, which starkly contrast with the rigid and regular objects commonly encountered in general object recognition, such as cars or chairs. Standard square convolution kernels have inherent limitations when capturing these free-form structures. To address this challenge, the current study employs deformable convolution (DCNv4)~\cite{wang2024dcnv4} to effectively represent irregular shapes. As illustrated in Figure \ref{fig:Difference}, unlike standard convolutions with fixed sampling locations, deformable convolutions learn a two-dimensional offset for each sampling point within the kernel. This mechanism allows the receptive field to dynamically adapt to the actual shape of the target, thereby enabling more accurate delineation of the boundaries and primary regions of the disease spots.

Figure \ref{fig:Shape} illustrates the detailed structure of the shape branch. First, the input feature descriptor is processed by DCNv4, after which a 1×1 convolutional layer generates a single-channel shape attention map. Notably, no activation function is applied at this stage (the Texture branch utilizes a similar approach, the details of which are omitted in subsequent sections). Consequently, the branch output—referred to as the “score”—can assume either positive or negative values. Positive scores indicate strong consistency with target features, whereas negative scores are associated with non-target or background features. Applying an activation function at this stage would nullify all negative values, thereby discarding approximately 50\% of the representational capacity. The subsequent fusion layer is designed to learn how to accommodate both positive and negative inputs, preserving the full range of the original scores and providing maximal flexibility for the final fusion process.

\subsubsection{Texture-Aware Branch (TextureBranch)}
In addition to shape features, crop diseases can be differentiated based on their unique surface textures. The Gabor filter~\cite{jain1991unsupervised}, a classical tool in texture analysis, is essentially a sinusoidal wave modulated by a Gaussian function that effectively responds to specific frequencies and orientations of texture information. Its response characteristics are notably analogous to the behavior of simple cells in the human primary visual cortex, thereby supporting its use within biologically inspired vision systems. While traditional methods typically employ a set of Gabor filters with fixed parameters to satisfy specific task requirements, the significant variability in texture features among different diseases necessitates a more adaptive approach. Accordingly, this study introduces a learnable Gabor filter set for the adaptive extraction of distinctive textures.

\begin{figure}[h]
\centering
\includegraphics[width=0.9\linewidth]{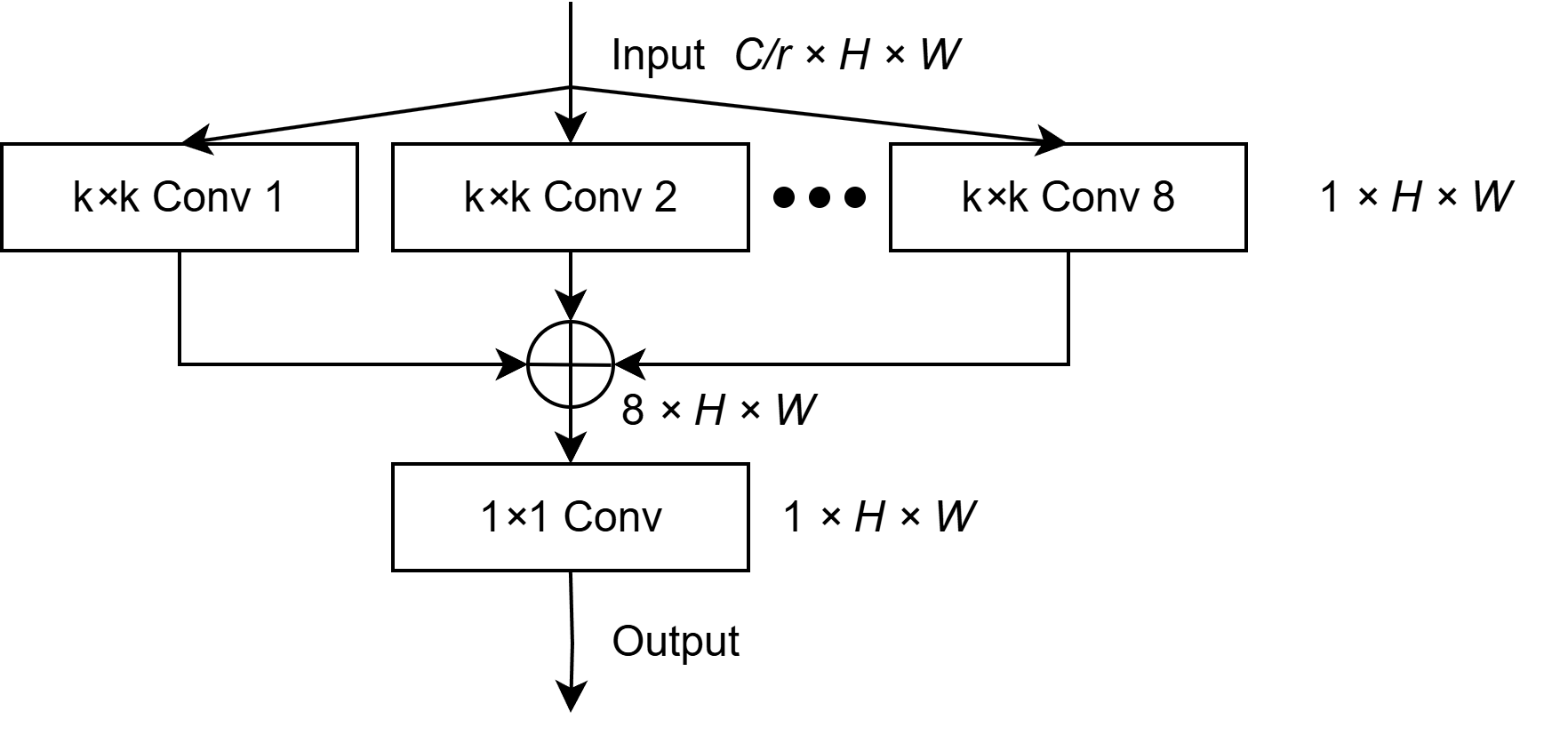}
\caption{\label{fig:Texture}TextureBranch}
\end{figure}

\begin{figure}[h]
\centering
\includegraphics[width=0.9\linewidth]{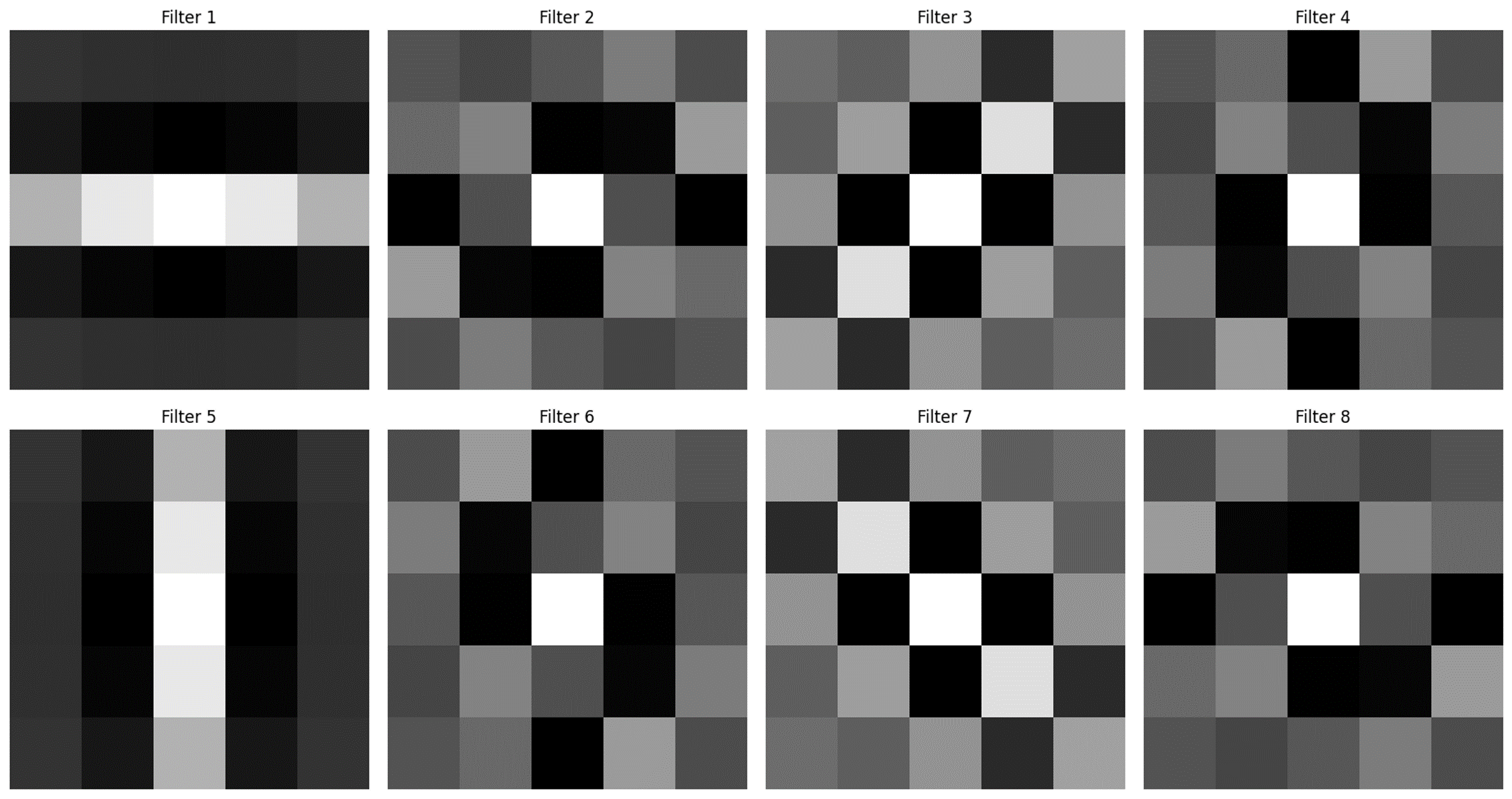}
\caption{\label{fig:GaborFilter}GaborFilter}
\end{figure}

Figure \ref{fig:Texture} details the architecture of the TextureBranch. A Gabor convolutional layer, composed of eight filters with diverse initial orientations and trainable weights, forms the foundation of this branch. Figure \ref{fig:GaborFilter} illustrates the scenario when the module hyperparameter is set to k = 5. During training, the network continuously fine-tunes these weights via backpropagation, enabling the filter set to progressively evolve into an "expert detector" that targets the specific pathological texture features present in the dataset. Finally, the outputs from all eight filters are concatenated and processed through a fusion convolution layer, ultimately generating a single-channel texture attention map.

\section{Experiment}
\subsection{Dataset}
This study employs the CCMT dataset~\cite{asenso2023ccmt}, an open collection of crop disease images released in 2023. As detailed in Table \ref{tab:ccmt}, the dataset comprises images of four major crops—cashew, cassava, maize, and tomato—collected in Ghana and Kenya. It includes 22 distinct categories that represent both healthy states and various disease conditions. The dataset contains 24,881 original images captured under diverse background and environmental conditions (as illustrated in Table \ref{tab:example}) and meticulously annotated by plant virology experts to ensure high data quality and diversity. For experimental evaluation, the dataset was partitioned into training, validation, and test sets in a 6:2:2 ratio.

\begin{longtable}[]{@{}
  >{\centering\arraybackslash}p{(\linewidth - 10\tabcolsep) * \real{0.1561}}
  >{\centering\arraybackslash}p{(\linewidth - 10\tabcolsep) * \real{0.1845}}
  >{\centering\arraybackslash}p{(\linewidth - 10\tabcolsep) * \real{0.1410}}
  >{\centering\arraybackslash}p{(\linewidth - 10\tabcolsep) * \real{0.1567}}
  >{\centering\arraybackslash}p{(\linewidth - 10\tabcolsep) * \real{0.2206}}
  >{\centering\arraybackslash}p{(\linewidth - 10\tabcolsep) * \real{0.1410}}@{}}
\caption{\label{tab:ccmt}CCMT Dataset}\\
\toprule\noalign{}
\multicolumn{6}{@{}>{\centering\arraybackslash}p{(\linewidth - 10\tabcolsep) * \real{1.0000} + 10\tabcolsep}@{}}{%
\begin{minipage}[b]{\linewidth}\centering
CCMT Dataset
\end{minipage}} \\
\midrule\noalign{}
\endhead
\bottomrule\noalign{}
\endlastfoot
\multicolumn{2}{@{}>{\centering\arraybackslash}p{(\linewidth - 10\tabcolsep) * \real{0.3406} + 2\tabcolsep}}{%
Class Labels} & Samples &
\multicolumn{2}{>{\centering\arraybackslash}p{(\linewidth - 10\tabcolsep) * \real{0.3773} + 2\tabcolsep}}{%
Class Labels} & Samples \\
\multirow{5}{=}{Cashew} & healthy & 1368 & \multirow{5}{=}{Cassava} &
healthy & 1193 \\
& anthracnose & 1729 & & bacterial\_blight & 2614 \\
& gumosis & 392 & & brown\_spot & 1481 \\
& leaf\_miner & 1378 & & green\_mite & 1015 \\
& red\_rust & 1682 & & mosaic & 1205 \\
\multirow{7}{=}{Maize} & healthy & 1041 & \multirow{5}{=}{Tomato} &
healthy & 500 \\
& fall\_armyworm & 1424 & & leaf\_blight & 1301 \\
& grasshoper & 3364 & & leaf\_curl & 518 \\
& leaf\_beetle & 4739 & & septoria\_leaf\_spot & 2343 \\
& leaf\_blight & 5029 & & verticulium\_wilt & 773 \\
& leaf\_spot & 5437 &
\multicolumn{3}{>{\centering\arraybackslash}p{(\linewidth - 10\tabcolsep) * \real{0.5183} + 4\tabcolsep}@{}}{%
\multirow{2}{=}{-}} \\
& streak\_virus & 5049 \\
\end{longtable}

\begin{longtable}[]{@{}
  >{\centering\arraybackslash}m{(\linewidth - 10\tabcolsep) * \real{0.1161}}
  >{\centering\arraybackslash}m{(\linewidth - 10\tabcolsep) * \real{0.1900}}
  >{\centering\arraybackslash}m{(\linewidth - 10\tabcolsep) * \real{0.1979}}
  >{\centering\arraybackslash}m{(\linewidth - 10\tabcolsep) * \real{0.1091}}
  >{\centering\arraybackslash}m{(\linewidth - 10\tabcolsep) * \real{0.1900}}
  >{\centering\arraybackslash}m{(\linewidth - 10\tabcolsep) * \real{0.1970}}@{}}
\caption{\label{tab:example}CCMT Dataset Example}\\
\toprule\noalign{}
\begin{minipage}[b]{\linewidth}\centering
\end{minipage} & \begin{minipage}[b]{\linewidth}\centering
anthracnose
\end{minipage} & \begin{minipage}[b]{\linewidth}\centering
gumosis
\end{minipage} & \begin{minipage}[b]{\linewidth}\centering
\end{minipage} & \begin{minipage}[b]{\linewidth}\centering
bacterial\_blight
\end{minipage} & \begin{minipage}[b]{\linewidth}\centering
brown\_spot
\end{minipage} \\
\midrule\noalign{}
\endhead
\bottomrule\noalign{}
\endlastfoot
Cashew &
\includegraphics[width=0.94488in,height=0.94488in]{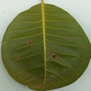} &
\includegraphics[width=0.94488in,height=0.94488in]{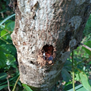} &
Cassava &
\includegraphics[width=0.94488in,height=0.94488in]{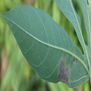} &
\includegraphics[width=0.94488in,height=0.94488in]{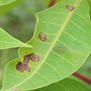} \\
& fall\_armyworm & leaf\_blight & & leaf\_blight & leaf\_curl \\
Maize &
\includegraphics[width=0.94488in,height=0.94488in]{Examples/5.png} &
\includegraphics[width=0.94488in,height=0.94488in]{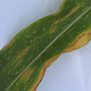} &
Tomato &
\includegraphics[width=0.94488in,height=0.94488in]{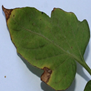} &
\includegraphics[width=0.94488in,height=0.94488in]{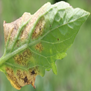} \\
\end{longtable}

\subsection{Experimental Configuration}
In this experiment, the hardware configuration comprises an Intel Core
i9-14900K central processing unit (CPU) and an NVIDIA RTX 4090D graphics
processing unit (GPU). The software environment features the PyTorch
2.6.0 deep learning framework, with the entire system deployed on the
Ubuntu 22.04.5 LTS operating system. Development and computational
acceleration are facilitated by Python 3.12.9 and CUDA 12.6.

To ensure a fair comparison, all models are trained using identical
hyperparameter settings. Each model undergoes training for 200 epochs
with a batch size of 256. The AdamW optimizer~\cite{loshchilov2017decoupled} is utilized, initialized
with a learning rate of \(3.0 \times 10^{- 3}\) and a weight decay
coefficient of 0.025. A Cosine Annealing Scheduler~\cite{loshchilov2016sgdr} is employed to
stabilize training and expedite convergence. During the initial five
epochs---a warm-up phase---the learning rate increases linearly from
\(1.0 \times 10^{- 4}\) to the designated initial value, before decaying
to a minimum of \(1.0 \times 10^{- 5}\) throughout the remainder of
training. The standard Cross-Entropy Loss function serves as the
training objective. In addition, Exponential Moving Average (EMA) is
incorporated with a decay coefficient of 0.99996 to enhance model
generalization, while an Adaptive Gradient Clipping (AGC) strategy~\cite{brock2021high} with
a threshold of 0.02 is applied to prevent gradient explosion. To enrich
the training data and mitigate overfitting, augmentation techniques
including ThreeAugment, RandAugment~\cite{cubuk2020randaugment}, and Random Erasing~\cite{zhong2020random} (with a probability of 0.25 for Random Erasing) are adopted. Moreover, all input images are uniformly resized to 224×224 pixels prior to network
ingestion.
\subsection{Evaluation Metrics}
To comprehensively evaluate the model's performance, this study utilizes
four key quantitative metrics: Top-1 Accuracy, F1-Score, Floating Point
Operations (FLOPs), and Model Parameters.

Top-1 Accuracy, a widely employed and intuitive metric in image
classification tasks, quantifies the proportion of instances in which
the category with the highest predicted probability matches the true
label. This metric directly reflects the model's overall ability to
accurately classify plant diseases. The calculation formula is presented
below:

\[Accuracy = \frac{\text{Number of Correct Predictions}}{\text{Total Number of Predictions}}\]

The F1-score is introduced as a supplementary evaluation metric to
address potential class imbalance issues commonly observed in plant
disease datasets, where the number of samples for certain diseases may
significantly exceed those of others. In such cases, relying solely on
accuracy can yield misleading evaluation results. Defined as the
harmonic mean of precision and recall, the F1-score provides a more
robust measure of a model\textquotesingle s overall performance across
classes, especially when dealing with imbalanced datasets. A higher
F1-score indicates a favorable balance between precision and recall. For
multi-class classification tasks, this study employs the macro-average
F1-score method, which involves calculating the F1-score for each class
and then averaging these values arithmetically, thereby ensuring that
each class is accorded equal importance during evaluation. The formula
for calculating the F1-score is as follows:

\[F1 = 2 \times \frac{Precision \times Recall}{Precision + Recall}\]

Precision is employed to evaluate the model's positive predictive
accuracy---that is, the proportion of samples predicted by the model as
belonging to a specific disease category that actually do. A higher
precision indicates that the model's predictions are more reliable,
reducing the likelihood of misclassifying healthy leaves or samples
exhibiting other diseases as the target disease. The precision is
computed using the formula:

\[Precision = \frac{TP}{TP + FP}\]

where TP (True Positive) represents the number of correctly predicted
positive samples, and FP (False Positive) represents the number of
samples mistakenly predicted as positive.

Recall, on the other hand, measures the model's capability to capture
all relevant cases, specifically reflecting the proportion of all actual
samples of a certain disease category that the model successfully
identifies. A higher recall demonstrates that the model can effectively
detect diseased samples, thereby minimizing false negatives---instances
where diseased leaves are not correctly identified. The recall is
calculated as:

\[Recall = \frac{TP}{TP + FN}\]

where FN (False Negative) represents the number of positive samples
incorrectly predicted as negative.

FLOPs (floating point operations) represent a key metric for assessing
the computational complexity of a model, as they quantify the amount of
processing required for a single forward pass. In scenarios where models
are deployed on edge devices, FLOPs have a direct impact on both
inference speed and energy consumption. A lower FLOPs value indicates
greater operational efficiency, thereby enhancing the
model\textquotesingle s suitability for real-time disease detection on
mobile or embedded platforms with constrained computational resources.

The parameter count, referring to the number of learnable elements
(e.g., weights and biases), serves as a critical measure of a model's
spatial complexity. This metric directly influences the storage
requirements and memory footprint of the model. Consequently, for edge
devices with limited storage, deploying lightweight models with fewer
parameters is essential for achieving efficient operation. This renders
the parameter count a fundamental criterion for evaluating the degree of
model compression.
\subsection{Experimental Setup and Results}
To validate the effectiveness of the STAM module and its synergistic
interaction with standard channel attention (SE), we designed five model
versions for comparative experiments as follows:

Baseline: Utilizes only the FNAS-Attention backbone network generated by
DeepMAD, without any additional attention modules.

+SE: Builds on the Baseline model by incorporating the SE channel
attention module---a high-performance and widely used channel attention
mechanism---into all MBConv modules.

+CBAM: Enhances the Baseline model by embedding the CBAM attention
module, which integrates both channel and spatial attention mechanisms,
into all MBConv modules.

+STAM: Augments the Baseline model by inserting the STAM module
selectively at the ends of stage 2 and stage 3.

+SE+STAM (Ours): Extends the +SE model by further introducing the STAM
module at strategic locations (specifically, at the end of stage 2 and
stage 3) to enhance functionality.

Additionally, we conducted two sets of experiments using MobileNetV3~\cite{howard2019searching} and MobileNetV4~\cite{pham2024mobilenetv4}---two mainstream lightweight models configured with the minimal capacity settings as officially released.

\begin{longtable}[]{@{}
  >{\centering\arraybackslash}p{(\linewidth - 8\tabcolsep) * \real{0.4036}}
  >{\centering\arraybackslash}p{(\linewidth - 8\tabcolsep) * \real{0.1454}}
  >{\centering\arraybackslash}p{(\linewidth - 8\tabcolsep) * \real{0.1537}}
  >{\centering\arraybackslash}p{(\linewidth - 8\tabcolsep) * \real{0.1537}}
  >{\centering\arraybackslash}p{(\linewidth - 8\tabcolsep) * \real{0.1437}}@{}}
\caption{\label{tab:result}Experimental Results}\\
\toprule\noalign{}
\begin{minipage}[b]{\linewidth}\centering
Network
\end{minipage} & \begin{minipage}[b]{\linewidth}\centering
Top-1 (\%)
\end{minipage} & \begin{minipage}[b]{\linewidth}\centering
Params (M)
\end{minipage} & \begin{minipage}[b]{\linewidth}\centering
FLOPs (M)
\end{minipage} & \begin{minipage}[b]{\linewidth}\centering
F1-Score
\end{minipage} \\
\midrule\noalign{}
\endhead
\bottomrule\noalign{}
\endlastfoot
MobileNetV3 & 88.26 & 1.698 & 67.8 & 88.21 \\
MobileNetV4 & 89.23 & 2.468 & 184.7 & 89.20 \\
\midrule
Baseline (NAS-Backbone) & 86.84 & 0.308 & 43.1 & 86.71 \\
\textbf{Baseline + STAM} & \textbf{87.41} & \textbf{0.332} & \textbf{50.8} & \textbf{87.29} \\
Baseline + SE & 88.03 & 0.378 & 43.5 & 87.94 \\
Baseline + CBAM & 87.20 & 0.327 & 44.1 & 87.06 \\
\textbf{Baseline + SE + STAM (Ours)} & \textbf{89.00} & \textbf{0.401} & \textbf{51.1} & \textbf{88.96} \\
\end{longtable}

The experimental results are presented in Table \ref{tab:result}. Compared with the baseline model, the addition of the STAM module alone increased the
Top-1 accuracy significantly---from 86.84\% to 87.41\%. This improvement
validates the efficacy of our disease feature-oriented STAM module in
capturing key visual information, outperforming the generic
spatial-channel mixed attention mechanism (CBAM, 87.20\%). Notably,
while the STAM module boosts accuracy substantially, it only adds 0.024
million parameters, demonstrating remarkable parameter efficiency

Our final model, which integrates both the SE and STAM modules
(+SE+STAM), achieved the highest accuracy of 89.00\%, markedly superior
to models employing a single attention module. This result strongly
suggests a synergistic effect between the SE channel attention and the
proposed STAM spatial attention. In this configuration, the SE module
emphasizes channel-level information enhancement, whereas the STAM
module precisely localizes and reinforces critical pathological feature
regions, enabling the model to process and interpret image information
more effectively.

Furthermore, when compared with the smallest versions of MobileNetV3
(1.698 million parameters and 67.8 million FLOPs) and MobileNetV4 (2.468
million parameters and 184.7 million FLOPs), our model demonstrates a
clear advantage in balancing efficiency and performance. Specifically,
our model utilizes only 0.401 million parameters and 51.1 million FLOPs
to achieve comparable or superior accuracy (89.00\% vs. 88.26\% vs.
89.23\%). In essence, our model employs only about one-sixth of the
parameters and less than one-third of the computational cost of
MobileNetV4, while attaining 99.7\% of its performance level. These
findings conclusively underscore the benefits of integrating a NAS
backbone with a dedicated attention module to construct an efficient,
lightweight model.

\section{Discussion}
\subsection{Synergistic Effects of STAM and Channel Attention}
Experimental results indicate that the integration of the STAM and SE modules yields optimal performance. This outcome is not coincidental but rather stems from the complementary nature of their functionalities. The SE module, functioning as a channel attention mechanism, primarily identifies the most critical feature channels by learning interdependencies among channels; it enhances channels that are information-rich while suppressing those that are noisy or redundant. In contrast, the STAM module operates as a spatial attention mechanism by pinpointing the most significant spatial locations on the feature map, thereby achieving precise localization of lesion regions through specialized analysis of shape and texture.  

Collectively, these modules establish an efficient "serial filtering" process: the SE module initially performs a "content screening" of the features, after which the STAM module executes "spatial localization" on the refined, high-quality features. This approach ensures that the computationally intensive STAM module concentrates exclusively on the most pertinent regions, thus optimizing computational efficiency. The pronounced synergistic effect elucidated here not only clarifies why straightforward module stacking falls short of the performance achieved by our carefully structured combination, but also highlights the robustness of our methodological design.

\subsection{The Importance of Strategic Arrangement of the Attention Module}
In this study, the STAM module is not arbitrarily integrated into the network; instead, it is meticulously designed and strategically positioned at the terminal stages of the network’s intermediate phase, specifically where the feature map sizes are 28×28 and 14×14. This approach is grounded in a comprehensive understanding of the hierarchical nature of convolutional neural network (CNN) features.

The shallow layers of the network are primarily responsible for capturing low-level features such as edges, corners, and color patches. Deploying complex modules like the STAM at this juncture is analogous to expecting a pathologist to derive meaningful conclusions from a single pixel—neither yielding the desired pathological patterns nor avoiding the introduction of noise. In contrast, the intermediate phase serves as a "golden zone" for feature extraction. Here, the network has already learned relatively complex local patterns, and the feature maps retain high spatial resolution along with rich semantic information. These maps clearly delineate key discriminative characteristics, including lesion shape and contour, color halos, and surface textures, thus providing an optimal platform for the STAM module's effective operation.

In the subsequent deep layers, the network undergoes extensive abstraction and semantic processing, leading to a significant reduction in spatial information; on a 7×7 feature map, the original visual forms and textural details are essentially obliterated. Consequently, the application of the STAM module in these layers not only fails to deliver the anticipated benefits but may also disrupt the learned high-level semantic representations.

Furthermore, positioning the STAM module at the end of a phase is a deliberate design choice. One may conceptualize a phase (for instance, the collection of blocks producing 28×28-sized features) as a "feature processing workshop," wherein multiple functional units iteratively refine, enhance, and combine features while maintaining spatial resolution. Only when the final block of the phase concludes its processing do the abstract features culminate in their most complete and mature form. Introducing the STAM module at this stage—as a "senior quality inspector"—ensures that decisions are based on the most information-rich and reliable feature maps, thereby facilitating the most accurate outcomes.

\section{Conclusion}
This paper presents a novel lightweight deep learning model, STA-Net, designed to address the challenges of high-precision plant disease classification on resource-constrained edge devices. The model is underpinned by a dual optimization strategy. First, a training-free neural architecture search (NAS) method is applied to generate an efficient computational backbone network. Second, a novel Shape-Texture Attention Module (STAM) is introduced, which incorporates domain-specific prior knowledge. Specifically, STAM enhances feature learning by decomposing the spatial attention mechanism into two distinct branches: a shape-aware branch that captures lesion contours and a texture-aware branch that analyzes pathological patterns.  

Extensive experiments conducted on the public CCMT dataset demonstrate the efficacy of the proposed approach. The final model achieves a validation accuracy of 88.96\% while maintaining an extremely low computational complexity of only 401,000 parameters and 51.1 million FLOPs. These results clearly surpass both the baseline model and models utilizing conventional attention mechanisms. Overall, this study offers a practical and efficient artificial intelligence solution for precision agriculture and validates a broader design principle: embedding domain-specific prior knowledge into the attention mechanism through a decoupled design can substantially enhance the performance of lightweight models in complex fine-grained recognition tasks.

\bibliographystyle{alpha}
\bibliography{Refer}

\end{document}